\documentclass{article}

\usepackage{PRIMEarxiv}

\usepackage{algorithm}
\usepackage{algorithmicx}
\usepackage{algpseudocode}

\usepackage{amsmath}
\usepackage{graphicx}
\usepackage{epstopdf}
\usepackage{enumitem}
\usepackage[utf8]{inputenc} 
\usepackage[T1]{fontenc}    
\usepackage{hyperref}       
\usepackage{url}            
\usepackage{booktabs}       
\usepackage{amsfonts}       
\usepackage{nicefrac}       
\usepackage{microtype}      
\usepackage{lipsum}
\usepackage{fancyhdr}       
\usepackage{graphicx}       
\graphicspath{{media/}}     
\usepackage{cite}

\title{Cyclic Graph Attentive Match Encoder (CGAME): 
A Novel Neural Network For OD Estimation 
}

\author{
  Guanzhou Li \\
  Tsinghua University \\
  Beijing, China\\
  \texttt{ligz19@mails.tsinghua.edu.cn} \\
  \And
  Yujing He \\
  Tsinghua University \\
  Beijing, China\\
  \texttt{hyj19@mails.tsinghua.edu.cn}\\
  \And
  Jianping Wu* \\
  Tsinghua University \\
  Beijing, China\\
  \texttt{jianpingwu@tsinghua.edu.cn} \\
}

\begin{document}
\maketitle

\begin{abstract}
Origin-Destination Estimation plays an important role in the era of Intelligent Transportation. Nevertheless, as a under-determined problem, OD estimation confronts many challenges from cross-space inference to non-convex, non-linear optimization. As a powerful nonlinear approximator, deep learning is an ideal data-driven method to provide a novel perspective for OD estimation. However, viewing multi-interval traffic counts as spatial-temporal inputs and OD matrix as heterogeneous graph-structured output, the existing neural network architecture is not suitable for the cross-space inference problem thus a new deep learning architecture is needed. We propose CGAME, short for cyclic graph attentive matching encoder, including bi-directional encoder-decoder networks and a novel graph matcher in the hidden layer with double-layer attention mechanism. It realizes effective information exchange between the forward networks and backward networks and establishes coupling relations across underlying feature space. The proposed model achieves state-of-the-art compared with baselines in the designed experiments and offers a paradigm for inference tasks across representation space.
\end{abstract}

\keywords{Origin-Destination estimation \and CGAME \and traffic demand estimation \and double-layer attention}

\section{Introduction}
%
%
%
%
Origin-Destination matrix (OD matrix), a significant concept in the transportation domain, depicts the demand between pair of spots. Traditionally, it serves as input in traffic simulations, and helps to assess urban construction layout then provide guidance on infrastructure development. With the development of Intelligent Transportation Systems (ITS), it plays an increasingly important role in traffic management and redistribution of shared vehicles. In the era of ITS, traffic management is no longer just about congestion reasoning of a individual intersection or road section, but requires coordinated control of the entire road network, in which the mastery of people's travel demand is indispensable. Besides, the efficiency improvement of shared mobility lies in optimizing the matching of supply and demand for travel as much as possible, and precise OD matrices will help operators better plan the routes of shared vehicles.

With regard to estimation method, Kalman filter (KF), Bayesian approach, Generalized Least Squares (GLS),  Maximum Likelihood (ML), and gradient-based techniques \cite{abrahamsson1998estimation} are common methods that have been used in the previous works. OD estimation can be regarded as the process of finding a solution of the following equation \cite{krishnakumari2019data}:

\begin{equation}
\label{eq:eq0}
    f_l(t)=\sum_{o,d,\tau}a_{od\tau}^{lt}d_{od}(\tau)
\end{equation}

where the target variable $d_{od}(\tau)$ denotes the number of trips from origin $o$ to destination $d$ departing at timestep $\tau$, and $f_l(t)$ gives traffic counts on link $l$ at timestep $t$. $a^{lt}_{od\tau}$ is a mapping tensor from $d_{od}(\tau)$ to $f_l(t)$ with 5 indices: $l$,$t$,$o$,$d$,$\tau$. OD estimation requires a derivation of high-dimensional parameter $a^{lt}_{od\tau}$ from low-dimensional observations $f_l(t)$, which makes it insolvable in a direct way\cite{behara2020novel, zhou2006dynamic}. Hence, except for a small amount of deliberately designed single-level models \cite{shen2012new}, the problem is often modeled as a hierarchical structure with upper-level estimating OD matrix and lower-level undertaking dynamic traffic assignment (DTA), and extra additional inputs or assumptions are often needed in the lower level. Most classical approaches based on this structure rely on specific traffic assignment models to establish the connection between OD flows to road flows. While in reality, the actual OD allocation is related to people's path selection behavior and many random influences during travel, making it much more complicated than an assignment matrix or simulation-based assignment, thereby affecting the accuracy of the allocation model and further affecting the accuracy of OD estimation. Apart from that, the classic approaches usually require some inputs of priori information, such as a priori OD matrix. These priori inputs are either based on the statistics of the historical data or on the output of the model at the previous moment. The bias of the a priori inputs tends to cause the accumulation of errors in the estimation results, thus making the estimation results less stable. As a model-free data-driven method, deep learning does not require analytical assignment model and priori inputs, which provides a novel perspective for the OD estimation, and its powerful estimation capability for nonlinearity and nonconvexity also makes it well suited to address this problem. However, the application of neural networks in OD estimation is relatively limited for the following reasons.

Regarding origins and destinations as points, each origin-destination pair can be viewed as an edge on a graph, and traffic counts depicts the spatial-temporal distribution of OD flows, then the OD estimation is essentially an approximation of quantities distributed on graph edges from spatial-temporal data. Building mappings across representation spaces is a challenging task, especially with the complicated topology of road networks and heterogeneous vehicular flow. 

To build a neural network framework for this challenging task, we make three assumptions. First, the population structure and urban layout will not change significantly in a short period. Second, people's statistical route choice habits can be reflected in long-term historical travel data, which means from the viewpoint that the assignment matrix depicts the probabilities of path selection and trip distribution, an explicit traffic assignment algorithm is not always necessary in OD estimation\cite{carrese2017dynamic}. Therefore, we launch an assignment matrix-free approach by applying an inverse encoder-decoder neural network to capture the mapping from OD flow to traffic counts on links in our work. Third, traffic counts at consecutive time intervals include information to approximate the quasi-dynamic OD matrix during the period \cite{marzano2018kalman}.

Following the discussions above, we take consecutive slices of traffic counts as input to estimate the quasi-dynamic OD matrix. To this end, we propose a novel neural network architecture, Cyclic Graph Attentive Matching Encoder (CGAME), bi-directional "Encoder-Decoder" networks coupled in embedding space with an attention mechanism. The main contributions of this study include:

\begin{enumerate}
\item We propose a novel end-to-end neural network approach for OD estimation with a forward network estimating demand and a backward network mining characteristics of trip assignments.
\item CGAME is designed as cyclic generation architecture integrated with double-layer attention in embedding space, enabling the networks of estimation and dynamic traffic assignment to exchange information from respective perspectives in the hidden layer.
\item In both experimental scenes, our method achieve highest accuracy compared with baselines
\item This architecture can be further extended to many other cross-space deduction scenarios.
\end{enumerate}
The rest of papers is organized as follows. Chapter II reviews related works on OD estimation. Chapter III give the problem statement in the perspective of deep learning. Chapter IV introduces detailed methodologies of CGAME. In Chapter V, we test our model in 2 scenarios and compare with baseline model. Conclusions are given in Chapter VI.

\section{Literature Review}
OD estimation is essentially the reverse engineering of traffic flow assignment, estimating the travel demand of people during a certain period of time from observable variables in the road network. Research on OD estimation originated in the 1980s \cite{van1978method,willumsen1979estimating} and has been extensively studied up to now, with two main differences between the various types of studies: data sources and estimating methods.

In terms of data sources, the earliest and most widely used data is the vehicle counts of the road, called traffic counts \cite{mussone2006od,antoniou2016towards,shao2015estimation}. As a basic data source in traffic, road traffic counts can be obtained from roadside detection units such as magnetic induction coils, camera video, and millimeter wave radar. However, since solving the OD matrix by single-moment traffic counts is under-determined, as given in equation \ref{eq:eq0}, some studies have observed continuous traffic counts over a certain period of time to obtain a quasi-dynamic OD matrix for this period, while some other studies supplement the missing information of the under-determined equations by other data sources \cite{lam2021origin}, such as GPS and mobile device \cite{iqbal2014development, ma2013deriving}, bluetooth MAC scanner \cite{behara2020novel}, automatic vehicle identification \cite{zhou2006dynamic, rao2018origin, cao2021day},  probe vehicular data \cite{cao2013bilevel}, smart card record \cite{munizaga2012estimation}, etc.
These data can reproduce partial routes of vehicles and reflect additional flow assignment information, making OD estimations more straightforward. Nevertheless, Since data upload, cleaning and alignment are labor-intensive and time-consuming processes, coupled with the limited infrastructure penetration of such data. For example, AVI has a maximum penetration rate of only ~40\%-80\%. Thanks to the popularity of smartphones, cell phone signaling data and GPS data have relatively high coverage. The positioning accuracy of cell phone signaling data depends on the spatial granularity of the base station and the temporal granularity of the use of the cell phone, and GPS data also has a location offset. The application of cell phone data also requires data cleaning, map matching and matching to traffic objects. Therefore, the subsequent processing and analysis process of these sources of data is crucial for the accuracy of the estimation. In addition, the accessibility of these data is not always the same, the damage rate of the device, and the usage rate of the cell phone always fluctuate over time, all of which pose challenges for the estimation and application of real-time OD. We would like to propose a deep learning architecture that can give real-time traffic demand data based on real-time traffic count data only.

The approaches of estimation are categorized into constrained optimizations, iterative state estimations, and gradient-based techniques.

Constrained optimizations refer to Maximum Entropy (ME), Maximum Likelihood Estimation (MLE), and Generalized Least Square (GLS). Maximum Entropy were first formally introduced by \cite{jaynes1957information} and then widely extended to other fields, and the method first appeared in OD estimation in the 1970s in the work of Zuylen and Willumsen. This method is concise and convenient to give in terms of formulas, but its probability-based nature does not guarantee that correct estimates will necessarily be obtained in this way, and it does not fully exploited the features embedded by historical data. Another easy formulaic approach is MLE \cite{ben1985alternative}, which estimates OD matrix by maximizing the likelihood of observations given a prior joint probability distribution. The main disadvantage of MLE is that it requires a prior distribution of traffic demand. Besides, factorial are needed to establish the mapping between traffic demand and the corresponding observation, which can be computationally intensive. To make the numerical calculations more solvable, \cite{aerde2003estimation} formulated a simplified version of MLE with the approximation of Stirling's formula. GLS is a kind of least squares on multiple variables, considering the covariance relationship between different variables, introduced first by Cascetta and Bell to minimize the errors between estimated OD and prior OD, simultaneously minimizing the errors between assigned traffic counts and measurements, subject to assignment equations or inequalities \cite{cascetta1984estimation,bell1991estimation}. Nevertheless, the temporal locality of data extraction in the original version might cause measurement noises, \cite{lin2003gls} et al. extended the GLS by adopting time windows. Apart from the individual application of these methods, some studies have obtained better results by the combination of various methods. For instance, Xie et al presented a combined form of ME and least squares, namely ME-LS. In general, this group of approach is theoretical and has developed earlier. However, the modeling approach inevitably relies on some assumptions and constraints of the corresponding model, which may lead to fluctuations in the estimation performance in some complicated scenarios.

Iterative state estimations mainly refer to Kalman filter and its variants. Kalman filter (KF), a classical algorithm developed by Rudolf E. Kalman, utilizes a series of observations over time to approach the true unknown variables iteratively and has widely extensive applications in various fields from the guidance of vehicles to traffic volume forecasting. \cite{ashok1993dynamic} first brought KF into OD estimation viewing the issue as an autoregressive process. Yet, the conventional Kalman filter is based on the assumption of linearity, while the generation of traffic demand and the assignment of traffic flow are both nonlinear processes, making the extended Kalman filter for nonlinear problems a natural and better choice for OD estimation problems \cite{chang1994recursive,chang1996estimation}. But iterative state estimation faces the risk of error accumulation with iterations. To ensure the stability and robustness of the results, \cite{marzano2018kalman} adopts the quasi-static assumption in which the states over a period of time are fed into the model simultaneously, which reduces the noise in the iterative process to a certain extent. For large-scale road networks, this method using matrix multiplication is sensitive to slight errors in the inputs, and the iterative approach causes more or less accumulation of errors, which makes the application of this method limited.

Gradient-based techniques include Simultaneous Perturbation Stochastic Approximation (SPSA) and neural networks. SPSA decouples the processes of target OD estimation and traffic assignment, and simultaneously optimizes OD demands and traffic counts through stochastic approximation algorithm \cite{balakrishna2008incorporating,cipriani2011gradient,balakrishna2007offline}. Nevertheless, there is a common problem of gradient-based methods in SPSA, that is they are sensitive to the initialization of parameters, the scale of variables, and the step-size of gradient descent. Specifically, SPSA may not converge well applying the same suite of parameter for different pairs of OD trips when the traffic demand of road networks is heterogeneous. To address this, \cite{tympakianaki2015c} improved the performance and robustness of SPSA by an extension of SPSA applying clustering strategies, namely c-SPSA. The c-SPSA and its variants are widely adopted in diverse advanced OD estimation methods with satisfactory results, thus it is intuitive that another gradient-based approach, neural network, can be applied to this problem. In comparison to the explicit gradient expression in SPSA, the neural network employs the automatic gradient back-propagation mechanism, which enables a more flexible model framework design and provides a more powerful estimation capability for this complex nonlinear issue. Some early works with neural networks are like \cite{gong1998estimating} that made use of Hopfield neural networks for obtaining the trip matrix, then \cite{krishnakumari2019data} availed neural networks to create a mapping from link flows and link speeds to trip productions and attractions, and combined the results with trip distributions based on N-shortest-path to estimate the OD matrix. \cite{cao2021day} designed a latency-constrained autoencoder to learn the denoised decomposed features of the OD matrix, etc. Finally, to extract the features of demand patterns better and alleviate the impact of the curse of dimensionality in large-scale networks, auxiliary methods like Principle Component Analysis (PCA)\cite{mussone2006od,lorenzo2013od,djukic2012efficient}, CP tensor decomposition\cite{ren2017efficient}, non-negative Tucker decomposition\cite{cao2021day} are exploited to reduce dimension in this problem.

\section{Problem Statement}
\subsection{Problem Definition}

The purpose of our research is to dynamically estimate OD matrix within a reference period given traffic counts on each link for consecutive sub-periods. We define a road network $G$ made up of $n_p$ spots $P_1, P_2, \cdots, P_{n_p}$, and $n_l$ links $l_1,l_2,\cdots, l_{n_l}$ connecting these spots. The traffic demands between these spots within a period $t$ are involved in OD matrix $\mathbf{D} \in \mathbb{R}^{n_p\times n_p}$, of which element $\mathbf{D}(i,j)$ represents the number of trips from spot $P_i$ to $P_j$. By dividing the period $t$ into $n_t$ sub-periods with fixed duration, labelled as $t_1, t_2, \cdots, t_{n_t}$, the element in the matrix of traffic counts $\mathbf{F} \in \mathbb{R} ^{n_{l}\times n_{t}}$, noted as $\mathbf{F}(i,j)$ means traffic counts of link $l_i$ at sub-period $t_j$. In this paper, we attempt to avail neural networks trained with historical data to estimate $\mathbf{D}$ with $\mathbf{F}$ as input.

\subsection{Overview}

\begin{figure}[t]
    \centering
    \includegraphics[width=0.8\textwidth]{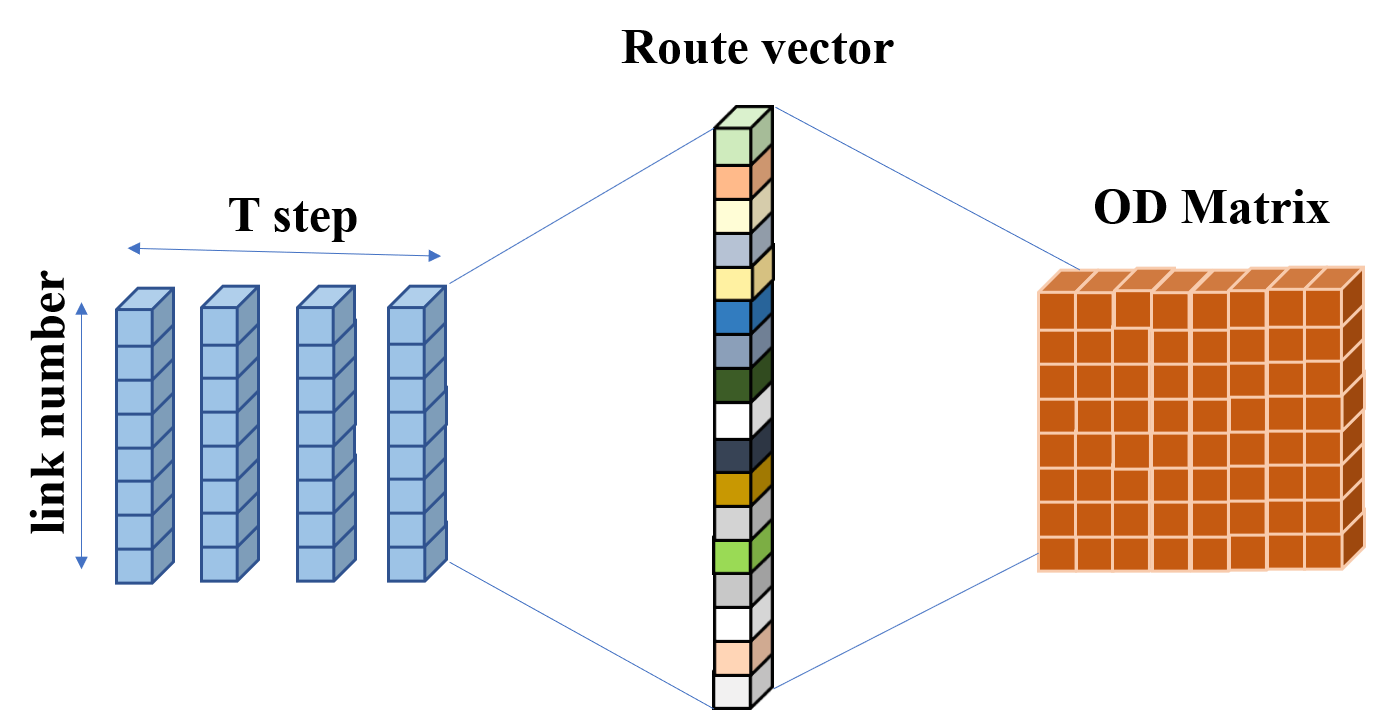}
    \caption{Mapping between Traffic Counts and OD Matrix by Route Distribution Vector}
    \label{fig:fig1}
\end{figure}

\begin{figure}[t]
    \centering
    \includegraphics[width=0.8\textwidth]{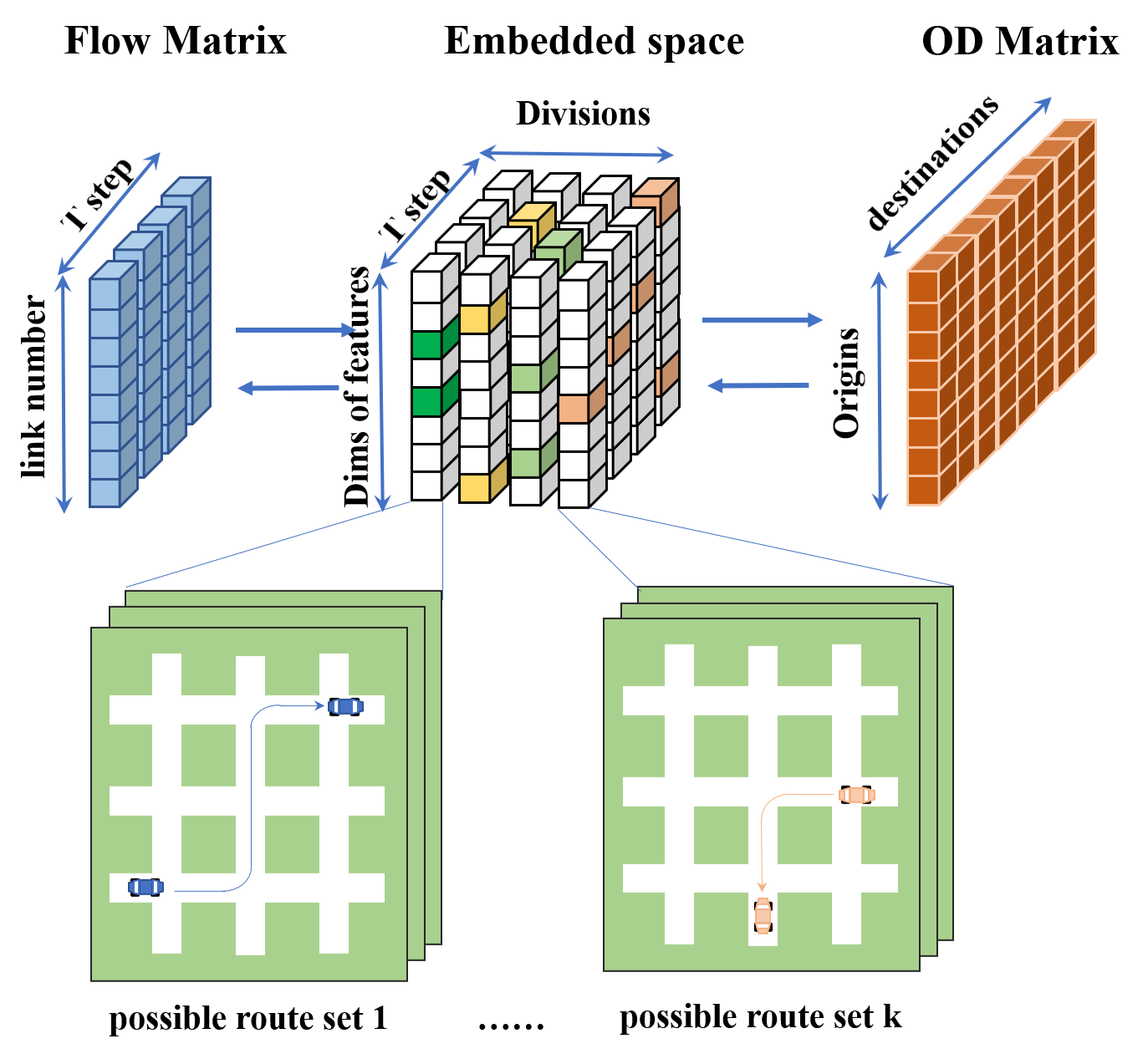}
    \caption{Mapping between Traffic Counts and OD Matrix by Embedded Feature Space}
    \label{fig:fig2}
\end{figure}

Trips are produced from one spot and sink into another spot via a sequence of links within a period of time, expressed as $P_{\varphi_o}\rightarrow l_{\alpha_1} \rightarrow l_{\alpha_2} \cdots \rightarrow l_{\alpha_{\eta}} \rightarrow P_{\varphi_d}$, where $P_{\varphi_o},P_{\varphi_d} \in \left\{P_1,P_2,\cdots,P_{n_p}\right\}$, and $l_{\alpha_1},\cdots,l_{\alpha_{\eta}} \in \left\{l_1,l_2,\cdots,l_{n_l}\right\}$. Taking the time into consideration, the section of a trip on the link $l_{\alpha}$ at sub-period $t_{\beta}$ can be marked as $(l_{\alpha},t_{\beta})$ with $\alpha$ and $\beta$ parameters, then the spatial-temporal route of a trip can be expressed as a sequence:
\begin{equation}
    r_k := \left[P_{\varphi_{o}},\left(l_{\alpha_{1}}, t_{\beta_{1}}\right),\left(l_{\alpha_{2}}, t_{\beta_{2}}\right), \ldots,\left(l_{\alpha_{\eta}}, t_{\beta_{\eta}}\right), P_{\varphi_{d}}\right]
\end{equation}
Define the number of trips along a specified route $r_k$ as $n_{r_k}$, then all trips in the period of time construct a vector 
\begin{equation}
    \mathbf{R}:=\left[n_{r_1},n_{r_2},\cdots,n_{r_k},\cdots,n_{r_{\gamma}}\right]
\end{equation}
with $\gamma$ routes totally. And each element of $\mathbf{F}$ and $\mathbf{D}$ can be expressed as:
\begin{equation}
\mathbf{F}(i, j)=\sum_{\left(l_{i}, t_{j}\right)\  in\  r_{k}}\   n_{r_{k}}
\end{equation}

\begin{equation}
    \mathbf{D}(i, j)=\sum_{\left(P_{\varphi_{O}}, P_{\varphi_{d}}\right)=\left(P_{i},P_{j}\right)} n_{r_{k}}
\end{equation}
respectively. From the equations we can see, the OD matrix reflects graph structure among spots, and traffic counts depict the spatial-temporal distributions of trips, features of graph structure and spatial-temporal distributions co-exist in the route vector $\mathbf{R}$ like two sides of a coin. The essential of OD estimation is to bridge these two sides by the route-distributed vector $\mathbf{R}$ as shown in figure \ref{fig:fig1}. In the direction from $\mathbf{D}$ to $\mathbf{F}$, dynamic traffic assignment generates a possible route distribution $\mathbf{R'}$, while in the inverse direction, OD estimation will give another possible route distribution $\mathbf{R''}$. To improve the accuracy of estimation, the co-occurrence probability of trips at time same position of $\mathbf{R'}$ and $\mathbf{R''}$ should be enhanced until $\mathbf{R'}$ closes to 
$\mathbf{R''}$. However, as $r_k$ combines spatial-temporal and graph-structured features, the dimensions of route-distributed vector $\gamma$ will increase to extremely high when the scale of road networks grows. In addition, as the sparsity of $\mathbf{R}$, the bi-direction approximations will be difficult and biased. To overcome the challenges, we compress vector $\mathbf{R}$ into embedding space with $n_s$ divisions and each division consists of $n_f$ features, each of which can be viewed as a subset of the whole route set $\{r_{\epsilon_1 },r_{\epsilon_2 },…,r_{\epsilon_v}\}$, then the matches between $\mathbf{R'}$ and $\mathbf{R''}$ are transferred to matches between vectors of features captured from forward and backward. And the $n_s$ divisions play the role as different channels shown as \ref{fig:fig2}.

Our proposed model, CGAME, adopts a bi-directional encoder-decoder architecture, with a midterm matching layer, namely graph matcher. It takes responsibility for seeking for the right matches to convey the gradient from one side to the other side precisely. To this end, the attention mechanism filters wrong matches and gives the correct matches higher passing rates. In each iteration, the encoder-decoder learns to message from one side to another by gradient propagation, and the graph matcher filters the correct messages to pass through. 

\section{Methodology}

\begin{figure*}[t]
    \centering
    \includegraphics[width=1.0\textwidth]{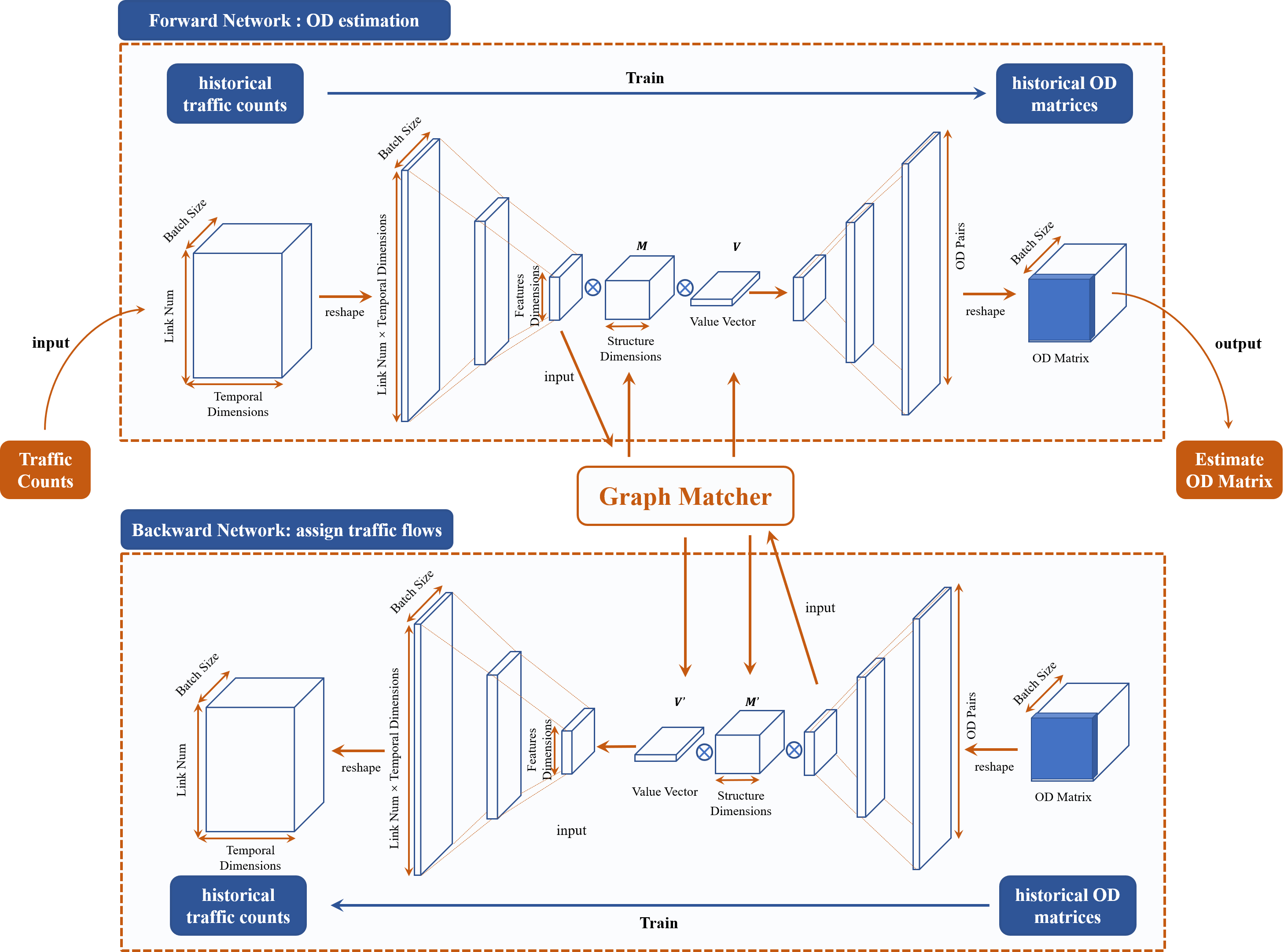}
    \caption{Framework of CGAME}
    \label{fig:fig3}
\end{figure*}

\subsection{Overall architecture}

CGAME consists of 3 components: forward encoder, backward encoder, and graph match layer \ref{fig:fig3}. Both forward and backward encoders avail encoder-decoder framework with a sharing midterm match layer, which captures the relations between the process of estimation represented by the forward network and the process of traffic assignment represented by the backward network. The historical traffic counts and historical OD matrices are fed to train the networks, and the forward network learns to approximate OD matrices. Wherein, the historical OD matrix can be obtained based on historical manual survey results, or can be obtained from license plate recognition data after cleaning and extraction, or trajectory reconstruction data from traffic flows, etc. When all these data sources lack, we can sample the closet results given by multiple other OD estimation models and utilize them to train the proposed model as true value.

For the convenience of expression, the notations in table \ref{tab:tab1} are used to represent relevant parameters and variables in CGAME.

\begin{table}[h]
    \caption{Notations}
    \centering
    \setlength{\tabcolsep}{12pt}
    \begin{tabular}{c|l}
    \toprule
    \textbf{Notation}    &   \textbf{Representation}\\
    \midrule
    $b$           &   batch size \\
    $h_x$         &   encoded vector in forward direction\\
    $g_x$         &   the results of $h_x$ after double-layer attention\\
    $h_y$         &   encoded vector in backward direction\\
    $g_y$         &   the results of $h_y$ afther double-layer attention\\
    $M$           &   structure matching matrix\\
    $V$           &   structure value matrix\\
    $j$           &   step of updating $M$ and $V$\\
    $p$           &   sub-step of updating $M$ in each step\\
    $\lambda_m$   &   discount factor in updating process of $M$\\
    $\lambda_v$   &   discount factor in updating process of $V$\\
    $n_f$         &   number of features in M\\
    $n_s$         &   number of structures in V\\
    $M_j$         &   M in the $j^{th}$ step\\
    $V_j$         &   V in the $j^{th}$ step\\
    \bottomrule
    \end{tabular}
    \label{tab:tab1}
\end{table}

\subsection{Encoder-Decoder Frameworks}

In the encoder-decoder architecture, we take multi-layer perceptrons (MLPs) to embed inputs into underlying features and decode the underlying features to outputs. The reason we adopt MLPs instead of the combination of graph convolutional networks (GCN) and recurrent neural networks which perform well in the capture of spatial-temporal features is to enable the graph matcher to obtain the underlying features from different directions better in a homogeneous encoding way. Besides, as a spectral graph convolution, GCN has the best performance with two layers in common, while route selection has characteristics of a wide range of multi-hops, hereby we need MLPs to connect all nodes to find the route distributions. Then the encoder part is built by two-layer MLPs expressed as:

\begin{equation}
    h_{x}=\texttt {LeakyReLU}\left(W_{2}\left(\texttt {LeakyReLU}\left(W_{1} x+b_{1}\right)\right)+b_{2}\right)
\end{equation}

where $W_1,W_2,b_1,b_2$ are learnable parameters in MLPs, $x$ is the input data, $h_x$ denotes the encoded features with shape of batch size $b$ times dims of features $n_f$. LeakyReLU, a frequently utilized activation layer, is an extensions of Rectified Linear Unit (ReLU).

With structure matching matrix $M \in \mathbf{R}^{n_f \times n_s}$ and structure vale matrix $V \in \mathbf{R}^{1\times n_s}$ given by the graph matcher, the attention operation is expressed by:

\begin{equation}
    g_{x}=mean_{n_{s}}(h_{x} \odot ||_{b} M \odot ||_{b, n_{f}} V)
\end{equation}

where $\odot$ represents Hadamard product operator, and a broadcast mechanism is availed in the process, $||_{b}$ denotes duplicating operation in the batch dimension, and $||_{b,n_f}$ denotes duplication in both batch and feature dimensions, $mean_{n_s}$ calculate mean value along the structure dimension. $h_x$, $g_x$ is the output of the encoder and the input of the decoder. More details about the representations and acquisition of $M$ and $V$ will be shown in the following part.

The decoder also includes double-layer MLPs, mapping the feature vector after attention $g_x$ to output:
\begin{equation}
    y=\texttt{LeakyReLU}(W_3(\texttt{LeakyReLU}(W_3g_x+b_3))+b_4)
\end{equation}

The similar encoder-decoder is applied in both directions.

\subsection{Graph Match Layer}

The structure matching matrix $M$ matches forward and inverse encoded features $h_x$ and $h_y$ and forms multiple matches from different perspectives, namely structures. The structure value matrix $V$ weighs structures by its quality of matching and gives different pass rates to various structures. In the beginning, $M$ and $V$ are both initialized as matrices filled with 1, on behalf of uniform structure with equal pass rates.

Every step update of $M$ consists of $n_s$ sub-steps and variation similarity along the batch dimension is calculated on each sub-step, with inherent logic that similar features from different sides vary in an analogous way. Specifically, the similarities between encoded features from the forward and backward encoders are decomposed into matches in different ranges, namely structures. In $p$-$th$ sub-step of $j$-$th$ step, $p$ batches of encoded features are concatenated to calculate structure in responding columns of structure matching matrix $M_j$:

\begin{equation}
\label{eq:eq1}
\begin{split}
    M_{j}(:,p) =(1-\lambda_{m}) \cdot M_{j}(:,p)+ \\
    \lambda_{m} \cdot \frac{\sum_{p\cdot b}\left(h_{x} \odot h_{y}\right)}{\sqrt{\sum_{p \cdot b}\left(h_{x} \odot h_{x}\right)} \sqrt{\sum_{p \cdot b}\left(h_{y} \odot h_{y}\right)}}
\end{split}
\end{equation}

where $M_{j}(:,p)$ denotes the $p$-$th$ column of structure matching matrix $M$. $\odot$ is Hadamard product, and $\sum_{p \cdot b}$ is the sum in the concatenated batch dimension with $p \cdot b$ data samples. Correspondingly, the $p$-$th$ column of structure value matrix $V$ is updated as:

\begin{equation}
\label{eq:eq2}
    V_{j}(p)=(1-\lambda_{m}) \cdot V_{j}(p)
\end{equation}

The structure value matrix $V$ measures the similarity between $h_y$ and $h_x'$, of which the latter denotes encoded feature vector $h_x$ operated by each structure, thus the update of $V$ is given by:

\begin{equation}
\label{eq:eq3}
    V_{j}=\frac{\sum_{n_{f}}\left( \left(\|_{n_{s}} h_{x}\right) \odot M_{j} \odot\left(\|_{n_{s}} h_{y}\right)\right)}{\sqrt{\sum_{n_{f}}\left(\left(\|_{n_{s}} h_{x}\right) \odot M_{j}\right)^2}\sqrt{\sum_{n_{f}}\left(\|_{n_{s}} h_{y}\right)^2}}+\lambda_{v} V_{j} 
\end{equation}

where $||_{n_s}$ duplicates encoded vectors along the structure dimension, and $\sum_{nf}(\cdot)$ is sum in the feature dim, $(||_{n_s} h_x) \odot M_j$ processes encoded feature vector $h_x$ to produce a sequence of $h_x'$ and $V_j$ essentially calculate cosine similarity between these $h_x'$s with $h_y$. Broadcasting mechanism is utilized in calculation. And the last term represents the discount weight value from last sub-step.

\begin{figure}[t]
    \centering
    \includegraphics[width=0.8\textwidth]{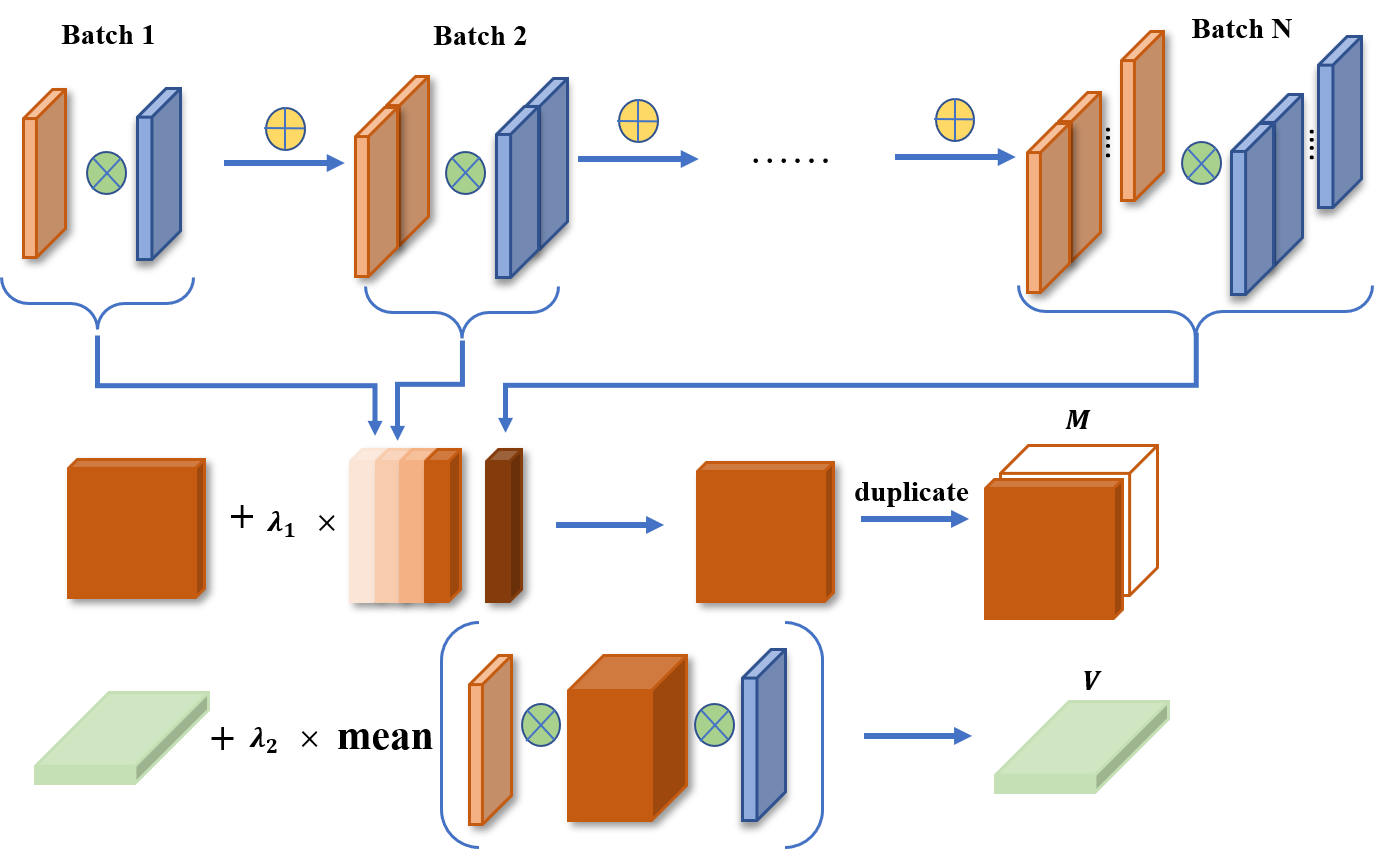}
    \caption{Graph Matcher}
    \label{fig:fig4}
\end{figure}

The pesudo-code of graph matcher is given in Algorithm \ref{alg:alg1}

\begin{algorithm}[h]
\caption{Graph Match Layer}\label{alg:alg1}
\begin{algorithmic}
  \State Input $h_{x,1},h_{x,2},\cdots,h_{x,n_s};h_{y,1},h_{y,2},\cdots,h_{y,n_s};\lambda_{m};\lambda_{v}$
  \State Initialize $M_0=1\in\mathbb{R}^{n_f\times n_s};V_0=1\in\mathbb{R}^{1\times n_s}$
  \For{$j=1,2,\cdots$}
  \For{$p=1,2,\cdots,n_{s}$}
  \State  $M_j(p)\leftarrow$ equation \ref{eq:eq1}
  \State $V_{j}(p) \leftarrow $ equation \ref{eq:eq2}
  \State $V_{j} \leftarrow$ equation \ref{eq:eq3}
  \EndFor
  \EndFor
  \State Return $M_j,V_j$\
\end{algorithmic}
\end{algorithm}

\subsection{Loss Function}

We compare two classical loss functions in deep learning, L1 Loss \ref{eq:eq4} and MSE Loss \ref{eq:eq5}, and the former can intuitively measure differences in vehicle number between estimation and true value, while the latter is able to restrict too large difference. According to the purpose of application, the loss function can choose from one of them.

\begin{equation}
\label{eq:eq4}
    \texttt{L1 Loss}=|y_t-\hat{y}_t|
\end{equation}
\begin{equation}
\label{eq:eq5}
    \texttt{MSE Loss}=\left(||y_t-\hat{y}_t||\right)^2
\end{equation}

\section{Experiments}
\subsection{Description}

We evaluate our model in 2 networks: one is a $6\times6$ grid network and the other is a realistic urban road network. And we simulate traffic networks with a widespread used simulator, SUMO. In simulations of both scenarios, vehicles are divided into 5 classes with different sizes and dynamic properties, including normal cars, slow cars, fast cars, buses, and trucks. Randomness is configured in the vehicle dynamics parameters, route selection, and simulation level.

\subsubsection{Scenario 1: Grid Networks}
\begin{figure}[t]
    \centering
    \includegraphics[width=0.8\textwidth]{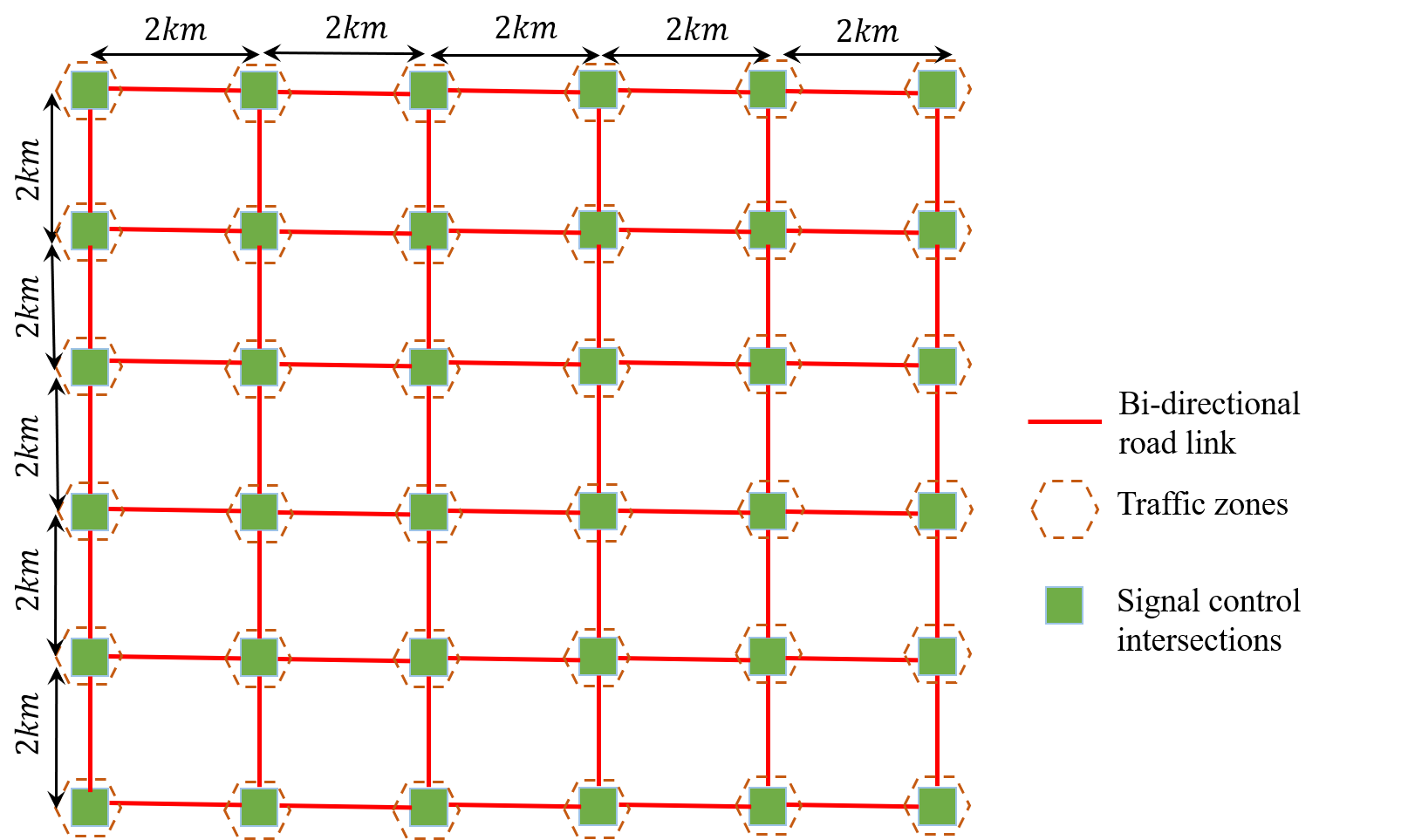}
    \caption{Grid Network}
    \label{fig:fig5}
\end{figure}

\begin{figure}[h]
    \centering
    \includegraphics[width=0.8\textwidth]{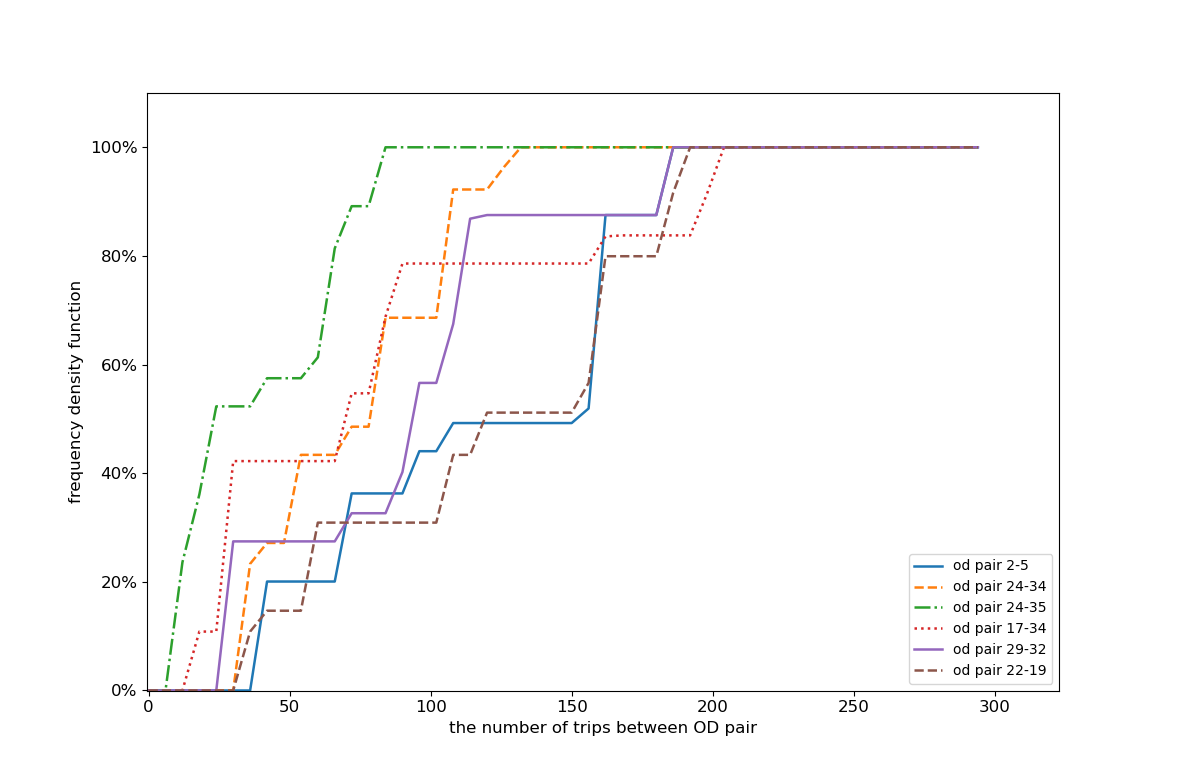}
    \caption{Frequency distribution function of trip number}
    \label{fig:fig6}
\end{figure}

\begin{figure}[t]
    \centering
    \includegraphics[width=0.8\textwidth]{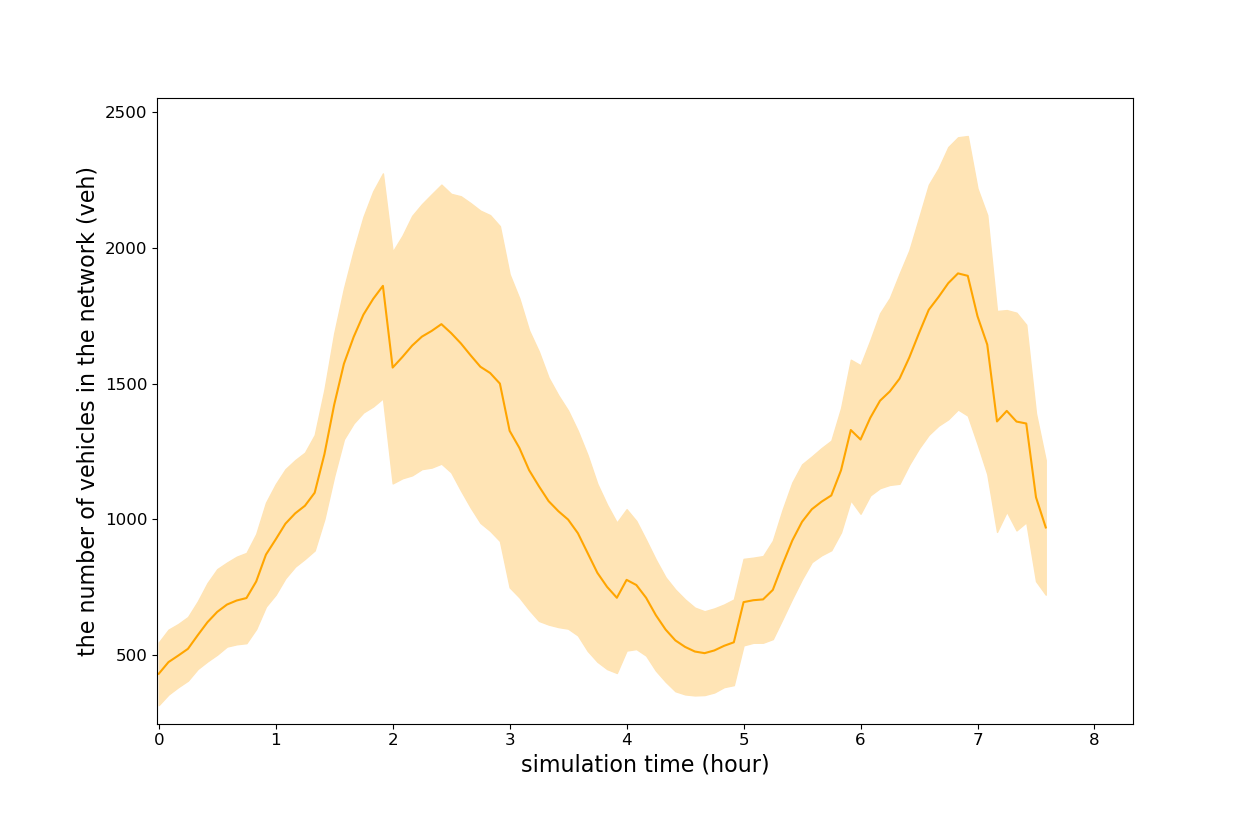}
    \caption{The number of vehicles in the network over time}
    \label{fig:fig7}
\end{figure}

A grid network as Fig \ref{fig:fig5} is studied in this scenario with $120$ 4-lane links and $32$ signaled intersections. The length of each link is $2km$ and the intersections include $16$ T-junctions and $16$ crossroads. There are $36$ traffic zones which vehicles depart from or arrive in, lying around the start point of each link. Then we construct twenty basic OD matrices as fundamental traffic demand patterns between traffic zones, and randomly sample one of these matrices to treat it as the true overall OD matrix after superimposing 20\% perturbations in every round of simulation. Specifically, figure \ref{fig:fig6} gives the frequency distribution curve of the trip number between randomly selected OD pairs in different simulation rounds. Throughout the simulation, all trips are stochastically selected time to departure during the entire simulation under a certain probability with two peaks. We count the number of trips from one spot to another per hour as real-time hourly OD matrices which fluctuate over time and estimate them by observations. The observations give the count of vehicles passing by each link every $5$ mins in the road network during the hour. The fluctuation of traffic counts over time can be shown as a curve of the total vehicle number in the road network in figure \ref{fig:fig7}. Besides, considering different principles of route selection by drivers, the route selection in the simulation is configured in $2$ ways. Given specified origin and destination, one is to choose paths with a relatively fixed probability distribution, the other is to adjust choices of routes according to the driver's perception of minimum travel time in real-time. The ratio of the former way is $70\%$ and the latter is $30\%$. All non-deliberately-detour routes between traffic zones are pre-designed as a route set from which drivers select their routes.

Each round of simulation involves 30000 simulated seconds, and we record the traffic counts and OD matrices from 2400 s to 27600 s for 7 hours in total. Thus 7 matrices of traffic counts with a shape of $12\times 120$ are leveraged to estimate 7 hourly OD matrices in every round. The matrix of traffic counts is made up of 12 consecutive 5-minute pieces of data measured from 120 links. The simulations run 1000 rounds totally and 7000 data samples are obtained, with 800 rounds of simulations taken as the training set and 200 rounds as the test set.

\subsubsection{Scenario 2: Realistic Urban Road Networks}
\begin{figure}[t]
    \centering
    \includegraphics[width=0.8\textwidth]{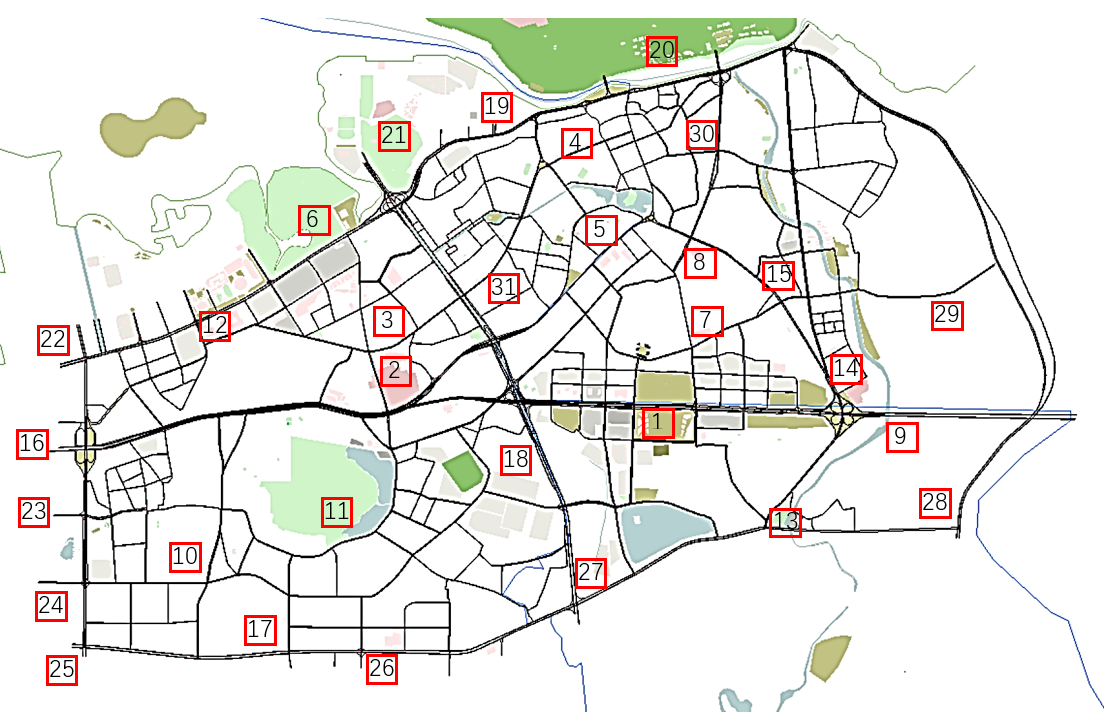}
    \caption{Haikou Network}
    \label{fig:fig8}
\end{figure}

A real urban road network from the downtown area with a size of $10km\times6km$ in Haikou, China is selected in the other scenario, including 2328 links and 1171 intersections\ref{fig:fig8}. Loops are placed on 359 links with high traffic volumes to meter traffic counts. Trips produce from and sink into 31 traffic zones involving multiple urban function partitions, and traffic generation and attraction of different urban functional areas show different changes over time in our simulation, like high traffic generation in residential areas and attraction in the business area in the morning rush hours, and reverse OD flows during evening rush hours. Other than that, as a city with a medium population, downtown Haikou covers several other types of typical travel demands. Specifically, for example, there are people going to the park for morning exercise or grocery shopping in the morning and people going to the night market in the evening, students returning home from school on Fridays, etc. And shopping malls become popular travel destinations on weekends. Based on that, we manually configure the weekly time-varying travel demands, and permute them with certain randomness in every round of simulation. 

The simulated duration within a day is from 5:20 a.m. to 10:20 p.m., divided into 6 periods: morning, morning peak, noon, afternoon, evening peak, and night. Considering the stabilizing process, the data we use in OD estimation is sampled from 6 a.m to 10 p.m. with data of hourly OD estimation and 12 pieces of 5-min traffic counts as described in the last scenario. The simulations run 400 rounds, producing 400-day data for 16h per day. After normalization and random permutation, 90\% of the data serves as the training set, and 10\% is taken as the test set. Note that the results are all de-normalized when testing.

\subsection{Evaluation metrics}
The metrics to evaluate the estimation results are given as below:
\begin{enumerate}[label=\alph*)]
    \item Root Mean Square Error (RMSE)
    \begin{equation}
        \texttt{RMSE}=\sqrt{\frac{1}{N} \sum_{i=1}^{N}\left(y_{i}-\hat{y}_{i}\right)^{2}}
    \end{equation}
    \item Mean Absolute Error (MAE)
    \begin{equation}
        \texttt{MAE}=\frac{1}{N} \sum_{i=1}^{N}\left|y_{i}-\hat{y}_{i}\right|
    \end{equation}
    \item Mean Absolute Percentage Error (MAPE)
    \begin{equation}
        \texttt{MAPE} = \frac{100\%}{N}\sum_{i=1}^{N}\frac{\left|y_{i}-\hat{y}_{i}\right|}{\left|y_{i}\right|}
    \end{equation}
    \item Coefficient of Determination ($R^2$)
    \begin{equation}
        R^{2}=1-\frac{\sum_{i=1}^{N}\left(y_{i}-\hat{y}_{i}\right)^{2}}{\sum_{i=1}^{N}\left(y_{i}-\bar{y}\right)^{2}}
    \end{equation}    
\end{enumerate}

where $N$ denotes the number of data samples from the test set, $y_i$ means the real value of the element in the OD matrix, and $\hat{y}_i$ is the corresponding estimated value by model, $\bar{y}_i$ is the mean of $y_i$

The RMSE and MAE measure the differences between the actual OD matrix and the estimation, while MAPE quantifies the relative deviation of the two. $R^2$ represents the correlation coefficient, depicting the correlation of data trends.

\subsection{Benchmark Model}
We select four baseline methods: two conventional methods and two neural network based methods. 
\begin{enumerate}
    \item Cluster-SPSA (c-SPSA), the approach that has been introduced in \cite{tympakianaki2015c}, following the idea in this study, the number of clustering kernels is selected as three.
    \item Extended Kalman Filter (EKF), the nonlinear extension for kalman filter which is widely-used in OD estimation.
    \item Classic spatial-temporal neural networks, in combination of graph convolutional networks (GCNs) and temporal convolutional networks (TCNs).
    \item Cyclic generative adversarial network (CycleGAN), originated in style transfer problem in computer vision, which has similar architecture with the proposed network with forward and backward GAN two encoders.
\end{enumerate}

\subsection{Results and Analysis}
\subsubsection{Measures of Performances}

\begin{table*}[h]
    \centering
    \caption{Measures of Performance}
    \begin{tabular}{c|cc|cc|cc|cc|cc}
    \toprule
             & \multicolumn{2}{c|}{c-SPSA} & \multicolumn{2}{c|}{EKF} &  \multicolumn{2}{c|} {GCN}     & \multicolumn{2}{c|}{CycleGAN} & \multicolumn{2}{c}{Proposed} \\ 
    \midrule
             & Grid         & Real     & Grid        & Real     & Grid    & Real & Grid           & Real       & Grid          & Real       \\ \cline{2-11} 
    RMSE     & 3.3751      & 6.5040     & 11.2133     & 13.0741    & 8.9487 & 8.0456      & 10.2721        & 11.3037       & 0.4566        & 4.0521       \\
    MAE      & 2.7255       & 4.9818     & 8.4872     & 10.1892    & 6.8846 & 6.0007      & 7.1281        & 8.4621       & 0.2951        & 3.4179       \\
    MAPE   & 14.83\%       & 22.64\%     & 50.29\%      & 64.63\%     & 28.12\%  & 20.66\%      & 34.98\%         & 30.79\%      & 2.78\%        & 15.09\%       \\
    R2       & *            & *          & *           & *          & *       & *      & *              & *            & 0.9963        & 0.5621       \\
    \bottomrule
    \end{tabular}
    \label{tab:tab2}
\end{table*}

Table \ref{tab:tab2} shows the performance of CGAME and baselines on 4 metrics in the grid and real network, where * means negative value, indicating poor performance in correlation $R^2$. The results show that CGAME obtains minimum RMSE, MAE, MAPE, and the highest accuracy in both the grid network and the realistic road network. The RMSE and MSE of CGAME decrease to 0.46 and 0.30 in the grid network, and around 4 in the real network. For most research on traffic demand, these errors are small enough that they do not affect the quality of data, which means that the OD data obtained by CGAME can be applied in many domains. This analysis is further convinced by relative error MAPE with $2.78\%$ and $15.09\%$ in two scenarios. Reasons for the decreased performance of the proposed model in the second scenario may include sparse rather than full traffic count measurements and increased complexity of the road network structure. Moreover, in order to simulate the traffic behavior of daily trips in the downtown to the greatest extent, various demands for different purposes and different types of route choices are taken into account. In general, the results of our model are significantly improved over the benchmarks in all test scenarios. The two neural network frameworks for comparison achieve similar results in the two test scenarios, and neither of them performs as satisfactorily on the OD estimation problem as on their original application domain, illustrating the inspiring perspectives of CGAME on introducing deep learning into traffic demand estimation problems. The poor performance of EKF is mainly due to the dependencies of a fixed assignment matrix and the accumulation of errors during the state iteration process under the condition that the traffic demand keeps fluctuating over time in our experiments. As a gradient-based optimization method, c-SPSA is employed in many of the advanced studies in demand estimation, and the method also has high accuracy as long as the traffic assignment model is accurate, and in our experiments it does achieve the best performance results among all benchmarks.

\begin{figure}[t]
    \centering
    \includegraphics[width=0.5\textwidth]{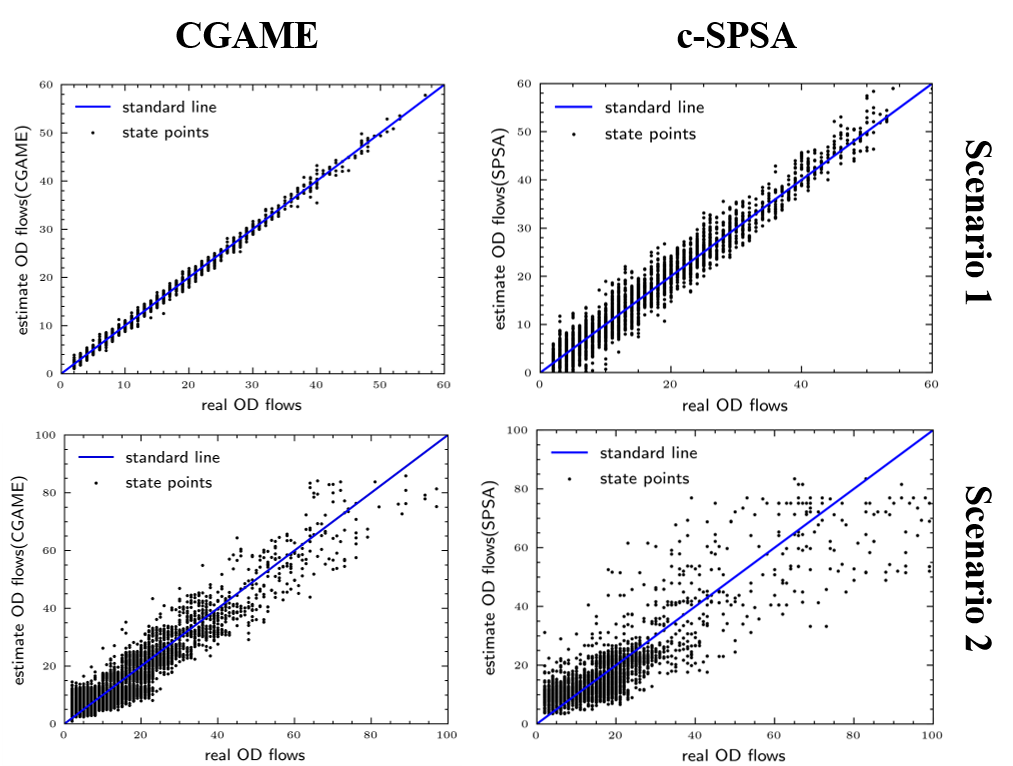}
    \caption{Comparisons of actual OD pairs and estimations}
    \label{fig:fig9}
\end{figure}

Figure\ref{fig:fig9} shows the estimation capability of GCAME and the best performing baseline, c-SPSA on traffic demand. we randomly sample 3000 pairs of actual OD values and their corresponding estimated values, namely state points. The standard line at $45^{\circ}$ inclination represents a 100\% accurate estimation, and the closer state points are to the standard line, the higher accuracy they have. In the case of a regular grid road network and the traffic counts of all roads are observable, CGAME shows extremely high accuracy, and there is no over-fitting in the experimental results, as the test set and training set are strictly separated and there are multiple randomness generation mechanisms in every round of simulations, as described in the first part of this chapter. 

\subsubsection{Comparison between actual OD matrices and estimations}
\begin{figure}[h]
    \centering
    \includegraphics[width=0.8\textwidth]{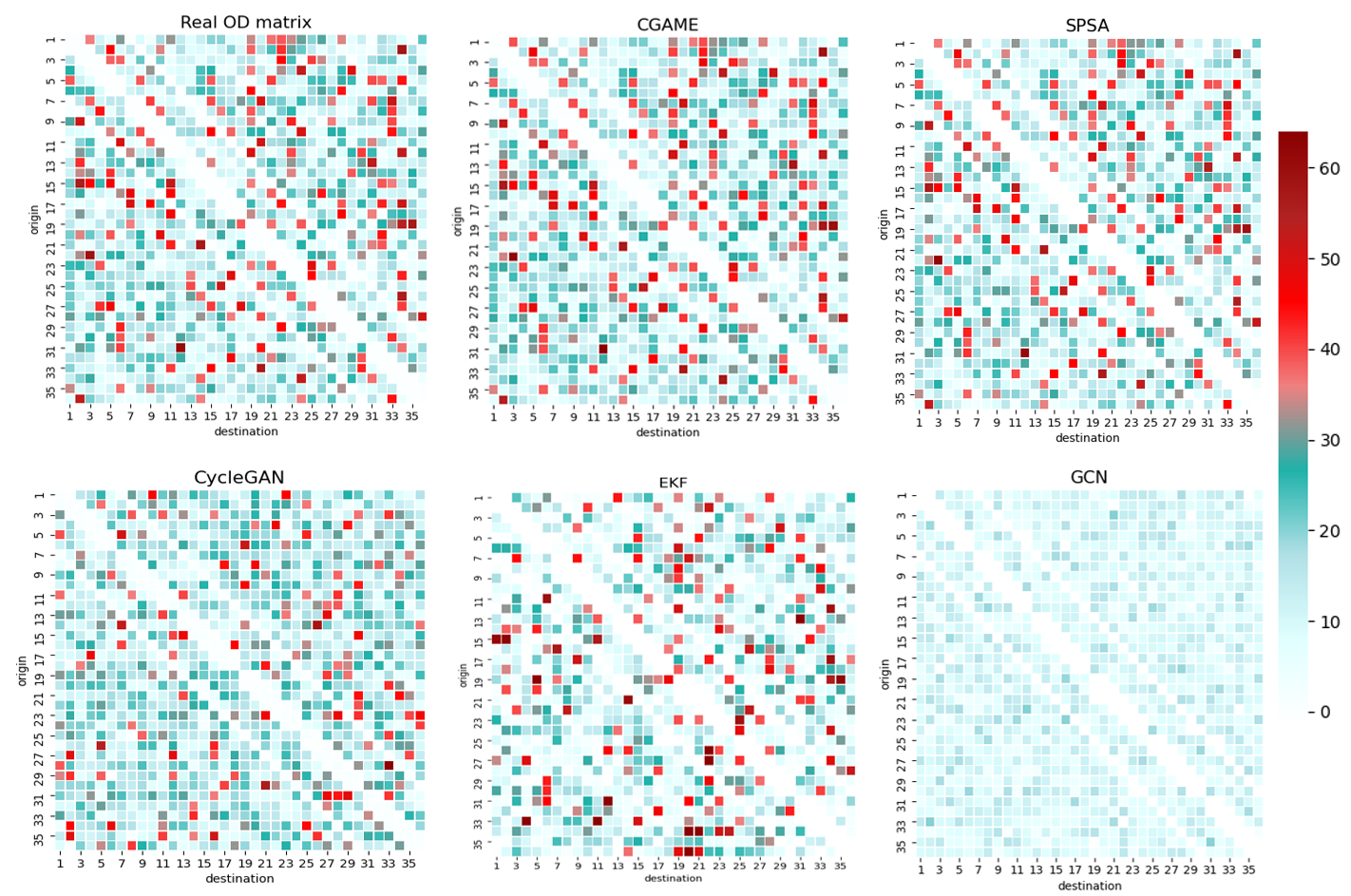}
    \caption{Thermodynamic diagram for grid network}
    \label{fig:fig10}
\end{figure}

\begin{figure}[h]
    \centering
    \includegraphics[width=0.8\textwidth]{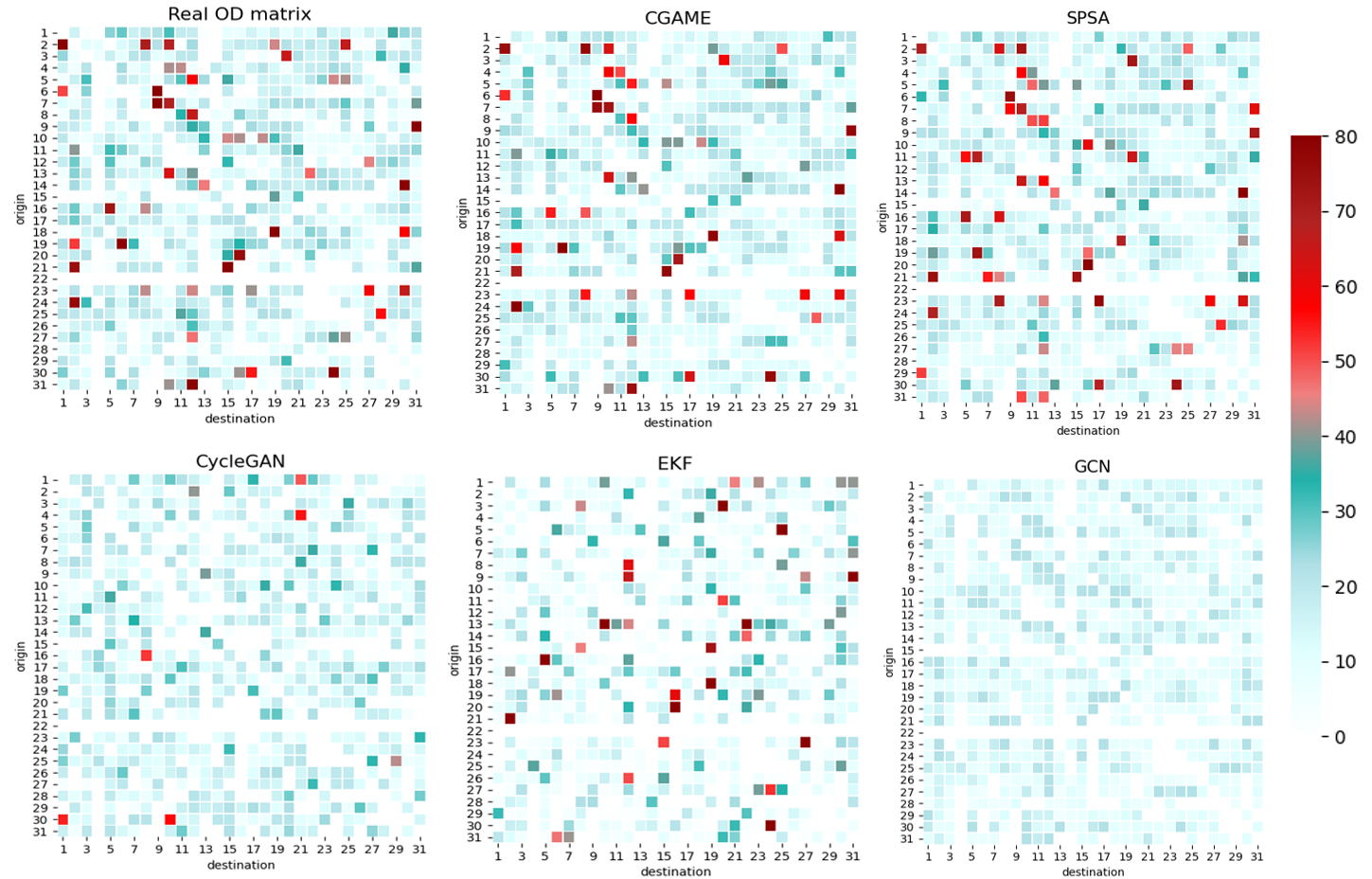}
    \caption{Thermodynamic diagram for real network}
    \label{fig:fig11}
\end{figure}

To present the results more intuitively, we present the thermodynamic diagram of actual and estimated OD as shown in figure\ref{fig:fig10} and figure\ref{fig:fig11}. Colors of pixels quantify the intensities of traffic demands between OD pairs. By specifying that OD flows cannot occur within a traffic zone itself or between adjacent traffic zones, which can be intuitively understood as OD trips being generated at one intersection and sinking at the next, hence there are three zero-value oblique lines around diagonal of OD matrix in figure \ref{fig:fig10}. Some other points, namely "heterogeneous points", like the read pixels, are noteworthy in the diagram. Their intensities of traffic demands are quite distinctive from their neighbor pixels', making it challenging for model to capture these points, while they are not negligible and even crucial in many cases. For example, the information of these prominent points plays a indispensable role in traffic management to equitably allocate traffic resources. The diagrams show that both our proposed model and c-SPSA are capable of identifying these outliers, and CycleGAN and EKF also produce these inhomogeneous points, but their distribution does not exactly match the distribution in the actual OD matrix. In the process of training CycleGAN, we find that even after fine-tuning, the loss of discriminator is still much smaller than that of the generator. Since the process of generation is a complex cross-spatial data transformation, it is hard to achieve a dynamic balance with the process of discrimination, which limit the implement of GAN in this issue. On the other hand, the images produced by GCN and TCN are smoother, and the major composition of low-demand OD pairs masks the capture of high-demand points, meaning this architecture of deep learning is fitter to extract overall rather than detailed distribution of traffic demand. In term of kalman filter, the iteration of states is composed of multi-step matrix multiplications, resulting in the elements in matrix easy to be too large or small collectively hence historical demand data is needed to calibrate the iteration process. As for our model, although table\ref{tab:tab2} shows decay in scenario two, it does not affect proposed model depicts the overall trend and detailed distribution of traffic demand, thus it is ideal for application in the field of OD estimation.

\subsubsection{Loss Curve}

\begin{figure}[t]
    \centering
    \includegraphics[width=0.8\textwidth]{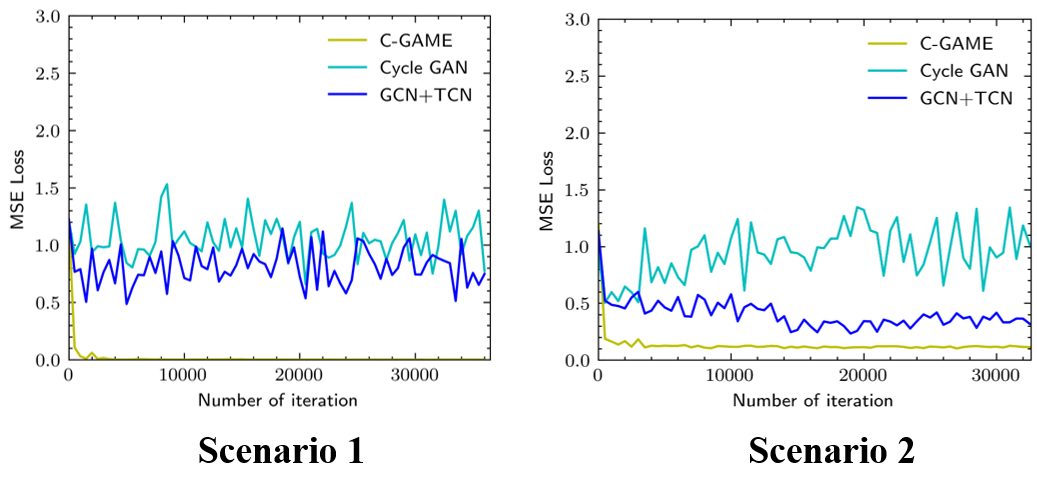}
    \caption{loss curves of neural networks}
    \label{fig:fig12}
\end{figure}

Figure \ref{fig:fig12} shows the training process of three neural networks, GCN, CycleGAN, and CGAME. The total training process includes 36000 iterations and the proposed model converges in around $1000$-$th$ iteration. The computation time on RTX 2060 are all around 40 minutes.

\section{Conclusion}
This research develops a novel neural network framework to address the OD estimation problem, involving a forward encoder for OD estimation, a backward encoder to extract the pattern of traffic assignment, and a double-layer attention mechanism exchange information between forward and backward learning. The framework is designed for cross-space inference problem like OD estimation with the introduction of concept about passing rate for the gradient propagation. And the proposed neural network framework can be generalized to many other analogous inference problems.

We evaluate the performance of our model with four baselines and reach the best performance. The test beds are configured as a $6 \times 6$ grid network and a real-size network in urban Haikou city with help of SUMO considering fidelity. In the experiment we find that, it is possible to estimate OD matrix accurately with partially consecutive observation of traffic counts. However, the complexity of the road network topology also affects the accuracy of OD estimation. Accurate traffic estimation in complex realistic urban road networks requires well-designed models and relatively complete historical data. In terms of model evaluation, the ability to accurately predict OD pairs with high demand is an important indicator for assessing the performance of an OD estimation model, and the proposed model exhibits satisfactory accuracy in the approximation of traffic demand.

Future research directions include studying the performance of the architecture with locally observed traffic demand data. We find that the performance of the model does not degrade significantly when there are missing values or random noises in the historical traffic demand. And by changing the structure of the two encoders in the architecture, such as using GCN, TCN, and CNN, the model can be extended to larger scale road networks. Besides, the architecture can be applied to many other areas, such as traffic flow prediction, traffic demand prediction, etc.

\section*{Acknowledgments}
The authors acknowledge support from the Center of High Performance Computing, Tsinghua University.

\bibliographystyle{unsrt}  
\bibliography{references}

\end{document}